\documentclass[10pt,twocolumn,letterpaper]{article}

\usepackage{cvpr}
\usepackage{times}
\usepackage{epsfig}
\usepackage[pagebackref=true,breaklinks=true,letterpaper=true,colorlinks,bookmarks=false]{hyperref}
\usepackage{graphicx}
\usepackage{amsmath}
\usepackage{amssymb}
\usepackage{cleveref}
\usepackage{textcomp}
\usepackage{subcaption}
\usepackage{enumitem}

\cvprfinalcopy % *** Uncomment this line for the final submission

\def\cvprPaperID{6} % *** Enter the CVPR Paper ID here

\ifcvprfinal\fi
\begin{document}

%%%%%%%%% TITLE
\title{Real-time Facial Surface Geometry from Monocular Video on Mobile GPUs}

\author{Yury Kartynnik \qquad Artsiom Ablavatski \qquad Ivan Grishchenko \qquad Matthias Grundmann\\
Google Research\\
1600 Amphitheatre Pkwy, Mountain View, CA 94043, USA\\
{\tt\small \{kartynnik, artsiom, igrishchenko, grundman\}@google.com}
}

\maketitle
%\thispagestyle{empty}

%%%%%%%%% ABSTRACT
\begin{abstract}
  We present an end-to-end neural network-based model for inferring an approximate 3D mesh representation of a human face from single camera input for AR applications. The relatively dense mesh model of 468 vertices is well-suited for face-based AR effects. The proposed model demonstrates super-realtime inference speed on mobile GPUs (100--1000+ FPS, depending on the device and model variant) and a high prediction quality that is comparable to the variance in manual annotations of the same image.
\end{abstract}

\section{Introduction}
The problem of predicting the facial geometry by aligning a facial mesh template, also called \emph{face alignment} or \emph{face registration}, has for a long time been a cornerstone of computer vision. It is commonly posed in terms of locating relatively few (typically 68) landmarks, or keypoints. These points either have distinct semantics of their own or participate in meaningful facial contours. We refer the reader to~\cite{Bulat} for a good review of related work on both the 2D and 3D face alignment problems.

An alternative approach is to estimate the pose, scale, and the parameters of a 3D morphable model (3DMM)~\cite{Blanz3DMM}. A 3DMM, such as BFM2017, the 2017 edition of the Basel Face Model~\cite{BFM2017}, is usually obtained through principal component analysis. The resulting mesh typically features many more points (around 50K in the case of BFM), but the range of possible predictions is limited by the linear manifold spanned by the PCA basis, which is in turn determined by the diversity of the set of faces captured for the model. As a concrete example, the BFM is seemingly incapable of reliably representing a face having exactly one eye closed.\footnote{Among the PCA expression basis components accounted for 97\% of the model variance, the one that controls eye closing is acting symmetrically on both eyes.}

\begin{figure}[tb]
    \centering
    \begin{subfigure}[b]{0.45\columnwidth}
        \includegraphics[width=\linewidth]{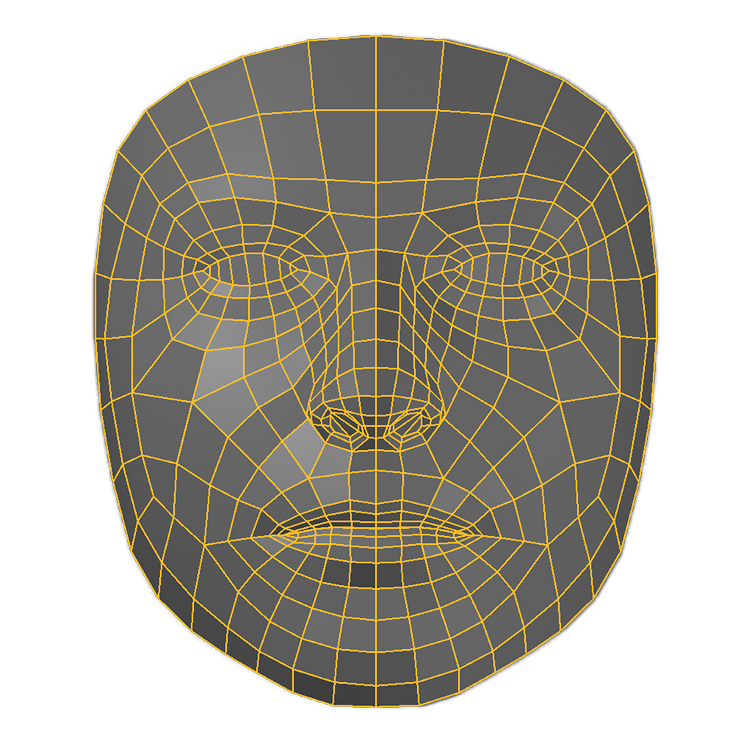}
        \caption{}
        \label{fig:topology}
    \end{subfigure}
    \begin{subfigure}[b]{0.45\columnwidth}
        \includegraphics[width=\linewidth]{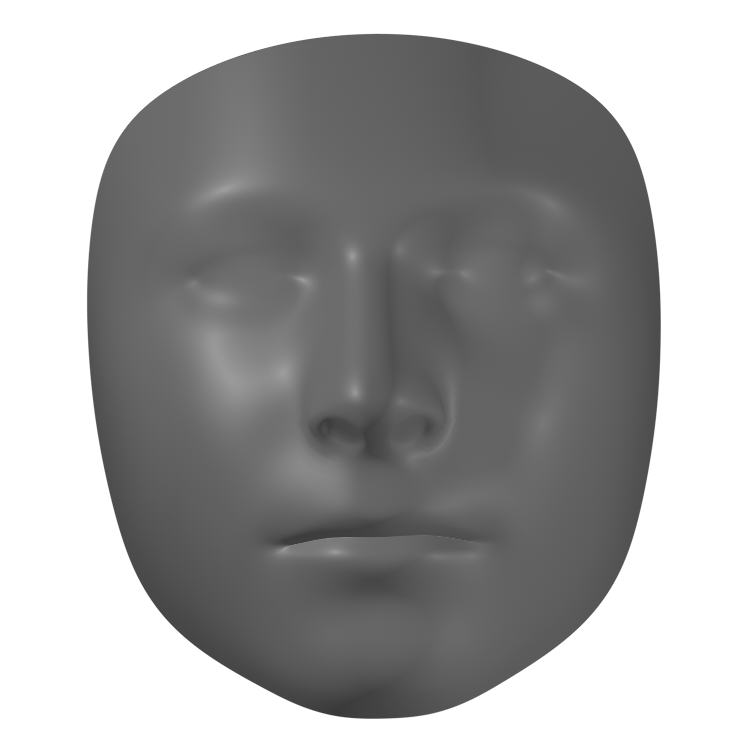}
        \caption{}
        \label{fig:smooth}
    \end{subfigure}
    \caption{\protect\centering The predicted mesh topology (a) and its 3-level Catmull-Clark subdivision (b)}
\end{figure}

We set forth a problem of estimating positions of the 3D mesh vertices with a neural network, treating each vertex as an independent landmark. The mesh topology is comprised of 468 points arranged in fixed quads (see \Cref{fig:topology}). The points have been manually selected in accordance with the supposed applications, such as expressive AR effects, virtual accessory and apparel try-on and makeup. The areas that are expected to have higher variability and higher importance in human perception have been allocated with higher point density. It allows to build a plausible smooth surface representation with the application of \eg Catmull-Clark subdivision~\cite{CatmullClark} (\Cref{fig:smooth}).

\begin{figure}[tb]
    \centering
    \begin{subfigure}[b]{0.45\columnwidth}
        \includegraphics[width=\linewidth]{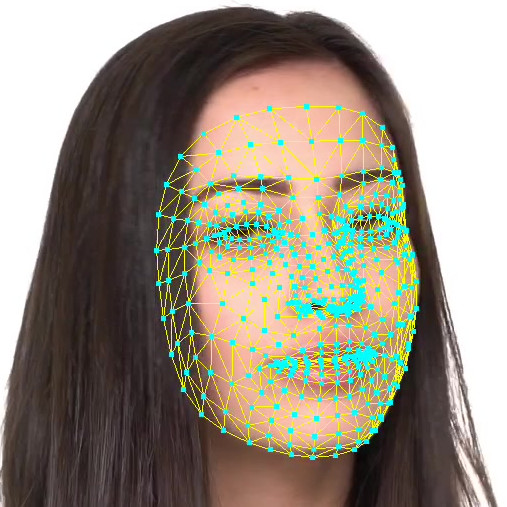}
    \end{subfigure}
    \begin{subfigure}[b]{0.45\columnwidth}
        \includegraphics[width=\linewidth]{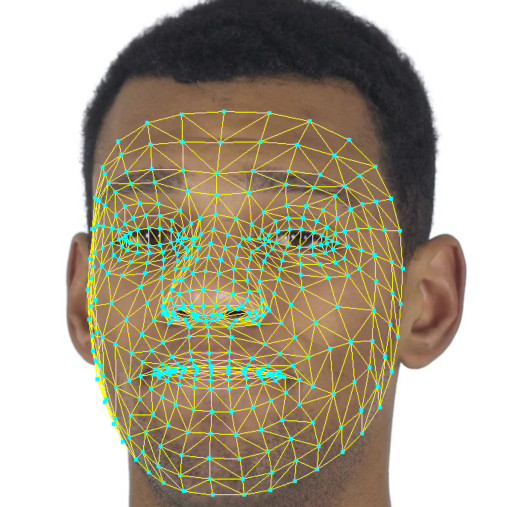}
    \end{subfigure}
    \caption{Face mesh prediction examples}
    \label{fig:examples}
\end{figure}

The input to the model is a frame (or, more generally, a stream of frames) of a single RGB camera---no depth sensor information is required. An example of the model output is presented in \Cref{fig:examples}. Our setup targets real-time mobile GPU inference, but we have also designed lighter versions of the model to address CPU inference on the mobile devices lacking proper GPU support. We will call the GPU-targeting model a ``full'' model, contrasting it to the ``lightest'' model tailored for CPU in the experiments.

\section{Image processing pipeline}

\label{sect:pipeline}
We organize the processing of an image as follows:
\begin{enumerate}[leftmargin=*,noitemsep,partopsep=0pt,topsep=0pt,parsep=0pt]
\setlength{\parskip}{0pt}
\setlength{\itemsep}{0pt plus 1pt}
    \item The whole frame from the camera input gets processed by a very lightweight face detector~\cite{BlazeFace} that produces face bounding rectangles and several landmarks (\eg eye centers, ear tragions, and nose tip). The landmarks are used to rotate a facial rectangle to align the line connecting the eye centers with the horizontal axis of the rectangle.
    \item A rectangle obtained in the previous step is cropped from the original image and resized so as to form the input to the mesh prediction neural network (ranging in size from 256$\times$256 pixels in the full model to 128$\times$128 in the smallest one). This model produces a vector of 3D landmark coordinates, which subsequently gets mapped back into the original image coordinate system. A distinct scalar network output (\emph{face flag}) produces the probability of the event that a reasonably aligned face is indeed present in the provided crop.
\end{enumerate}

We have adopted the policy that the $x$- and $y$-coordinates of the vertices correspond to the point locations in the 2D plane as given by the image pixel coordinates. The $z$-coordinates are interpreted as the depth relative to a reference plane passing through the mesh's center of mass. They are re-scaled so that a fixed aspect ratio is maintained between the span of $x$-coordinates and the span of $z$-coordinates, \ie a face that is scaled to half its size has its depth range (nearest to farthest) scaled down by the same multiplier.

When used on video input in the face tracking mode, a good facial crop is available from the previous frame prediction and the usage of the face detector is redundant. In this scenario, it is only used on the first frame and in the rare events of re-acquisition (after the probability predicted by the face flag falls below the appropriate threshold).

It should be noted that with this setup, the second network receives the inputs with faces reasonably centered and aligned. We argue that this allows to save some model representational capacity that could otherwise be spent on handling the cases with substantial rotation and translation. In particular, we could reduce the amount of related augmentations while gaining prediction quality.

\section{Dataset, annotation, and training}
In our training, we rely on a globally sourced dataset of around 30K in-the-wild mobile camera photos taken from a wide variety of sensors in changing lighting conditions. During training, we further augment the dataset with standard cropping and image processing primitives, and also a few specialized ones: modelling camera sensor noise~\cite{DorfCurves} and applying a randomized non-linear parametric transformation to the image intensity histogram (the latter helps simulating marginal lighting conditions).

Obtaining the ground truth for the 468 3D mesh points is a labor-intensive and highly ambiguous task. Instead of manually annotating the points one by one, we employ the following iterative procedure.
\begin{enumerate}[leftmargin=*,noitemsep,partopsep=0pt,topsep=0pt,parsep=0pt]
\setlength{\parskip}{0pt}
\setlength{\itemsep}{0pt plus 1pt}
    \item Train an initial model using the following two sources of supervision:
    \vspace{-0.5\topsep}
    \begin{itemize}[leftmargin=*]
    \setlength{\parskip}{0pt}
    \setlength{\itemsep}{0pt plus 1pt}
        \item Synthetic renderings of a 3DMM over the facial rectangles of real-world photos (as opposed to \eg solid backgrounds, to avoid overfitting on them). The ground truth vertex coordinates are thus immediately available from a predefined correspondence between the 468 mesh points and a subset of 3DMM vertices.
        \item 2D landmarks corresponding to a small subset of the mesh vertices participating in a set of semantic contours (see \Cref{fig:contours_demo}) annotated over the actual ``in-the-wild'' dataset. The landmarks are predicted as a separate output at the end of a dedicated network branch, introduced with the intent to share the intermediate face representations between the 2D and 3D paths.
    \end{itemize}

    \begin{figure}[t]
        \centering
        \includegraphics[width=0.35\linewidth]{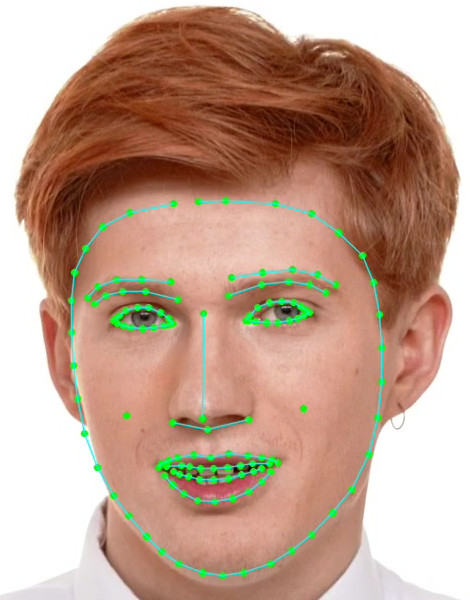}
        \caption{\protect\centering The 2D semantic contours used during the initial bootstrap process}
        \label{fig:contours_demo}
    \end{figure}

    After this first model had been trained, up to 30\% of the images in our dataset had predictions suitable for refinement in the subsequent step.
    
    \item Iteratively refine the $x$- and $y$-coordinates bootstrapped by applying the most up-to-date model to the images, filtering out those suitable for such refinement (\ie where the prediction error is tolerable). Fast annotation refinement is enabled by a ``brush'' instrument with adjustable radius that lets a whole range of points to be moved at once. The amount of movement is exponentially decreasing with the distance along the mesh edges from the pivot vertex under the mouse cursor. This allows annotators to adjust substantial area displacements with large ``strokes'' before local refinements, while preserving the mesh surface smoothness. We note that the $z$-coordinates are left intact; the only source of supervision for them being the synthetic 3D rendering outlined above. Despite the depth predictions being thus not metrically accurate, in our experience the resulting meshes are visually plausible enough to \eg drive realistic 3D texture renderings over the face or align 3D objects rendered as part of virtual accessory try-on experience.
\end{enumerate}

\section{Model architecture}
For the mesh prediction model, we use a custom but fairly straightforward residual neural network architecture. We use more aggressive subsampling in the early layers of the network and dedicate most of the computation to its shallow part. 

Thus, the neurons’ receptive fields start covering large areas of the input image relatively early.
When such a receptive field reaches the image boundary, its relative location in the input image becomes implicitly available for the model to rely on (due to convolution padding). Consequently, the neurons for the deeper layers are likely to differentiate between \eg mouth-relevant and eye-relevant features.

The model is able to complete a face that is slighly occluded or crossing the image boundary. This leads us to a conclusion that a high-level and low-dimensional mesh representation is built by the model that is turned into coordinates only in the last few layers of the network.

\section{Filtering for temporal consistency in video}
Since our model is operating on a single-frame level, the only information that gets passed between frames is the rotated facial bounding rectangle (and whether or not it should be re-evaluated with face detector). Because of the inconsistencies in pixel-level image representations of faces across subsequent video frames (due to small affine transforms of the view, head pose change, lighting variation, as well as different kinds of camera sensor noise~\cite{DorfCurves}), this leads to human-noticeable fluctuations, or \emph{temporal jitter}, in the trajectories of individual landmarks (although the entire mesh as a surface is less affected by this phenomenon).

We propose to address this issue by employing a one-dimensional temporal filter applied independently to each predicted landmark coordinate. As the primary application of our proposed pipeline is visually appealing rendering, we draw inspiration from human-computer interaction methods, specifically the 1 Euro filter~\cite{1Euro}. The main premise of 1 Euro and related filters is that in the trade-off between noise reduction and phase lag elimination, humans prefer the former (\ie stabilization) when the parameters are virtually not changing and the latter (\ie avoiding the lag) when the rate of change is high. Our filter maintains a fixed rolling window of a few timestamped samples for velocity estimations, which are adjusted by the face size to accommodate for face scale changes in a video stream. Using this filter leads to human-appealing prediction sequences on videos without visible jitter.

\section{Results}
We use mean absolute distance (MAD) between the predictions and the ground truth vertex locations, normalized by interocular distance (IOD), defined as the distance between the eye centers (estimated as midpoints of eye corner connecting segments to avoid gaze direction dependence). This normalization is aimed at avoiding factoring in the scale of the face. As the $z$-coordinates are obtained exclusively from the synthetic supervision, we report the 2D-only errors, but 3D inter-ocular distance is used to account for possible yaw head rotations.

To quantify the ambiguity of the problem and obtain a baseline for our metric, we have given the task to annotate a set of 58 images to each of 11 trained annotators and computed IOD-normalized mean absolute distance between their annotations of the same image. The estimated IOD MAD error was 2.56\%.

We present the evaluation results on a geographically diverse evaluation set of 1.7K images. The speed estimations are based on the TensorFlow Lite GPU framework~\cite{TFLiteGPU}.

% The data was taken from go/tflite-gpu-perf without time spent copying from/to CPU.
\begin{table}[ht]
\centering
\begin{tabular}{|l|c|c|c|c|}
\hline
Model (input) & IOD & Time, ms & Time, ms \\
 & MAD & (iPhone XS) & (Pixel 3) \\ \hline
Full (256$\times$256) & 3.96\% & 2.5 & 7.4 \\ \hline
Light (128$\times$128) & 5.15\% & 1 & 3.4 \\ \hline
Lightest (128$\times$128) & 5.29\% & 0.7 & 2.6 \\ \hline
\end{tabular}
\vskip 1ex
\caption{Model performance characteristics}
\label{tbl:inference_speed}
\end{table}

\begin{figure}[b]
    \centering
    \includegraphics[width=0.45\linewidth]{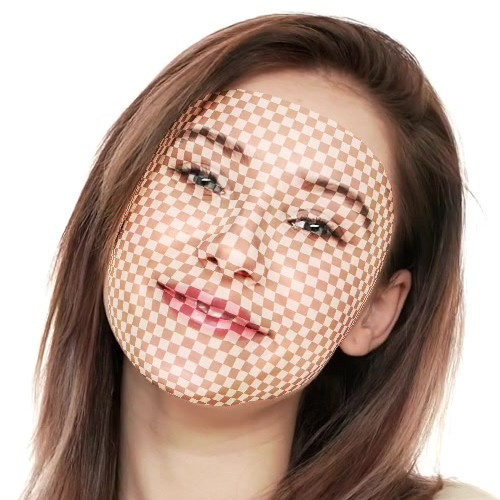}
    \includegraphics[width=0.45\linewidth]{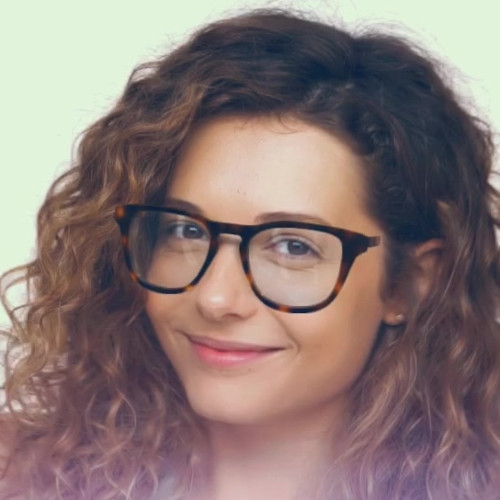}
    \caption{\protect \centering Application examples: Facial texture painting and AR object rendering (glasses)}
    \label{fig:effects}
\end{figure}

The technology described in this paper is driving major AR self-expression applications and AR developer APIs on mobile phones. \Cref{fig:effects} presents two examples of numerous rendering effects enabled by it.

{\small
\bibliographystyle{ieee_fullname}
\bibliography{meshbib}
}

\end{document}